\documentclass[10pt,twocolumn,letterpaper]{article}

\usepackage{cvpr}
\usepackage{times}
\usepackage{epsfig}
\usepackage{graphicx}
\usepackage{amsmath}
\usepackage{amssymb}
\usepackage{subcaption}
\usepackage{enumitem}

\usepackage[pagebackref=true,breaklinks=true,letterpaper=true,colorlinks,bookmarks=false]{hyperref}

\cvprfinalcopy 

\def\cvprPaperID{1} 

\ifcvprfinal\fi
\begin{document}

\title{Convolutions on Spherical Images}

\author{Marc Eder \quad \quad \quad Jan-Michael Frahm \\
University of North Carolina at Chapel Hill\\
{\tt\small \{meder, jmf\}@cs.unc.edu}
}

\maketitle

\begin{abstract}
Applying convolutional neural networks to spherical images requires particular considerations. We look to the millennia of work on cartographic map projections to provide the tools to define an optimal representation of spherical images for the convolution operation. We propose a representation for deep spherical image inference based on the icosahedral Snyder equal-area (ISEA) projection, a projection onto a geodesic grid, and show that it vastly exceeds the state-of-the-art for convolution on spherical images, improving semantic segmentation results by $12.6\%$.
\end{abstract}

\section{Introduction and Related Work}
Omnidirectional imaging is becoming increasingly popular thanks to the proliferation of consumer-grade $360^{\circ}$ cameras and the benefit of the wide field-of-view to a number of applications. At the same time, deep learning using convolutional neural networks (CNNs) has never been a more widely-used tool for computer vision tasks. Yet, applying CNNs to spherical images is not quite as trivial as simply training a vanilla CNN on spherical projections onto an image plane. Any planar representation of a sphere necessarily contains some degree of content deformation. This violates convolution's translation equivariance, which requires the convolution kernel and the signal to be uniformly discretized. To address the unique problems of convolution on spherical images, several methods have been proposed. 

\begin{figure}[ht]
    \centering
    \includegraphics[height=3.5cm] {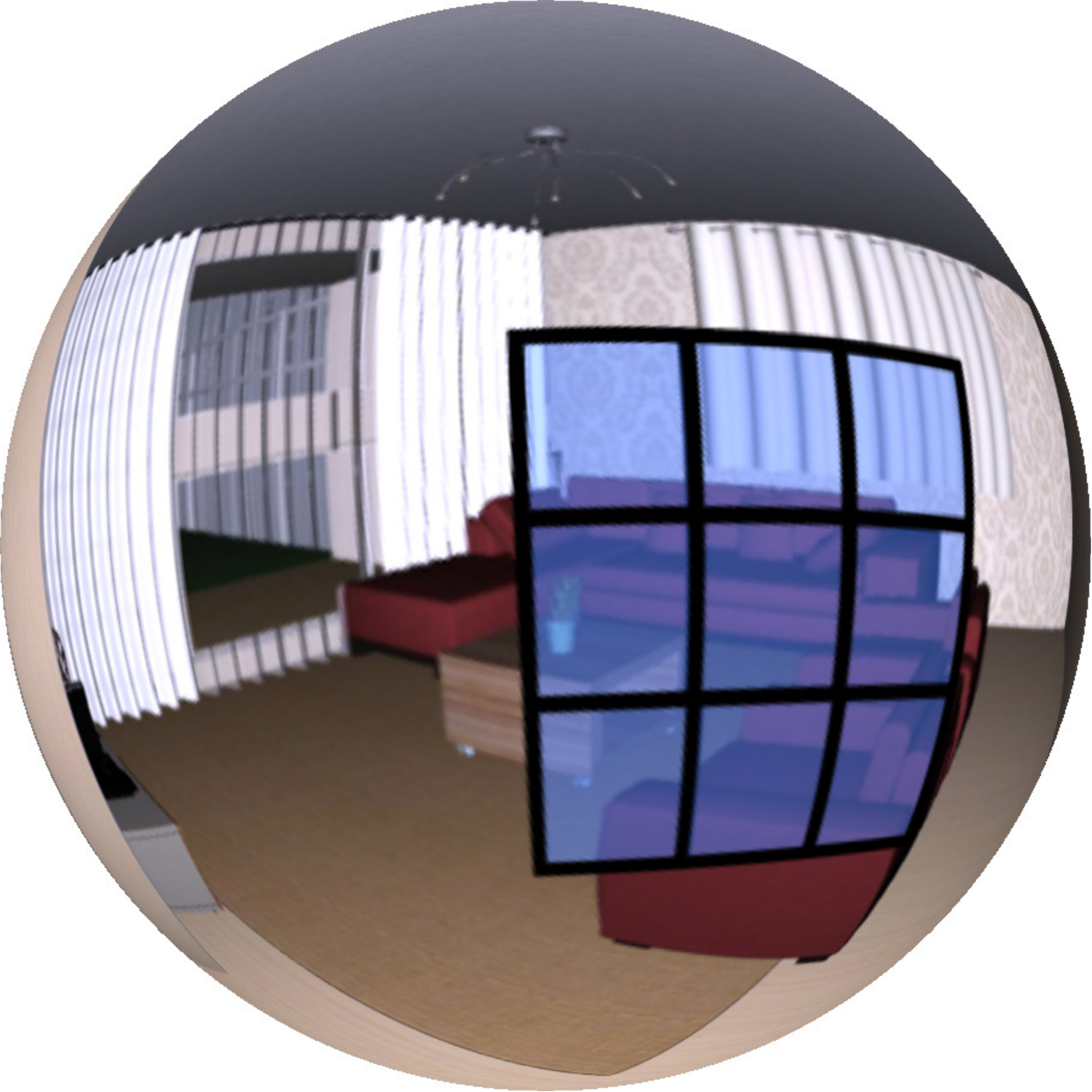}
\caption{We show that convolving on a planar geodesic approximation to a sphere drastically improves results for the dense prediction task of semantic segmentation.}
\label{fig:teaser}
\end{figure}

Su and Grauman \cite{su2017learning} train a CNN to transfer center-perspective-projection-trained models to the equirectangular domain using a learnable adaptive convolutional kernel to match the outputs. Observation that certain regions of cube-map faces contain less distortion than others, Xiong and Grauman \cite{xiong2018snap} develop a content-aware rotation of the spherical data that ``snaps-to'' a view where the relevant information has minimal distortion. Zioulis \etal \cite{zioulis2018omnidepth} explore the use of rectangular filter-banks, rather than square filters, to horizontally expand the network's receptive field to account for distortion in equirectangular images. The spherical convolution derived by Cohen \etal \cite{cohen2017convolutional, cohen2018spherical} uses a generalized formulation of convolution to filter rotations of the feature maps rather than translations. Finally, work from Coors \etal \cite{coors2018spherenet} and Tateno \etal \cite{tateno2018distortion} develop a location-dependent transformation of the planar convolutional kernel using a gnomonic (rectilinear) projection to adapt to the spatial distortion in spherical images.

We take a slightly different approach than these previous works. Rather than changing the tools of convolution, we look to find a better representation of the data. We first analyze the convolution operation to identify the important properties that must be considered when applying convolutions to spherical images and use this to determine a more optimal way to represent the data.
We resolve that a planar geodesic approximation to the sphere is a more optimal representation and show that its use yields state-of-the-art performance for dense semantic segmentation.

\section{Cartographic Map Projections}
First, we briefly review cartographic map projection properties and explain their relevance to this problem. 

\begin{figure*}[ht]
    \centering
    \begin{subfigure}[b]{0.24\textwidth}
        \centering
        \includegraphics[height=2.8cm, trim={0.1cm 0.1cm 0.1cm 0.1cm}, clip] {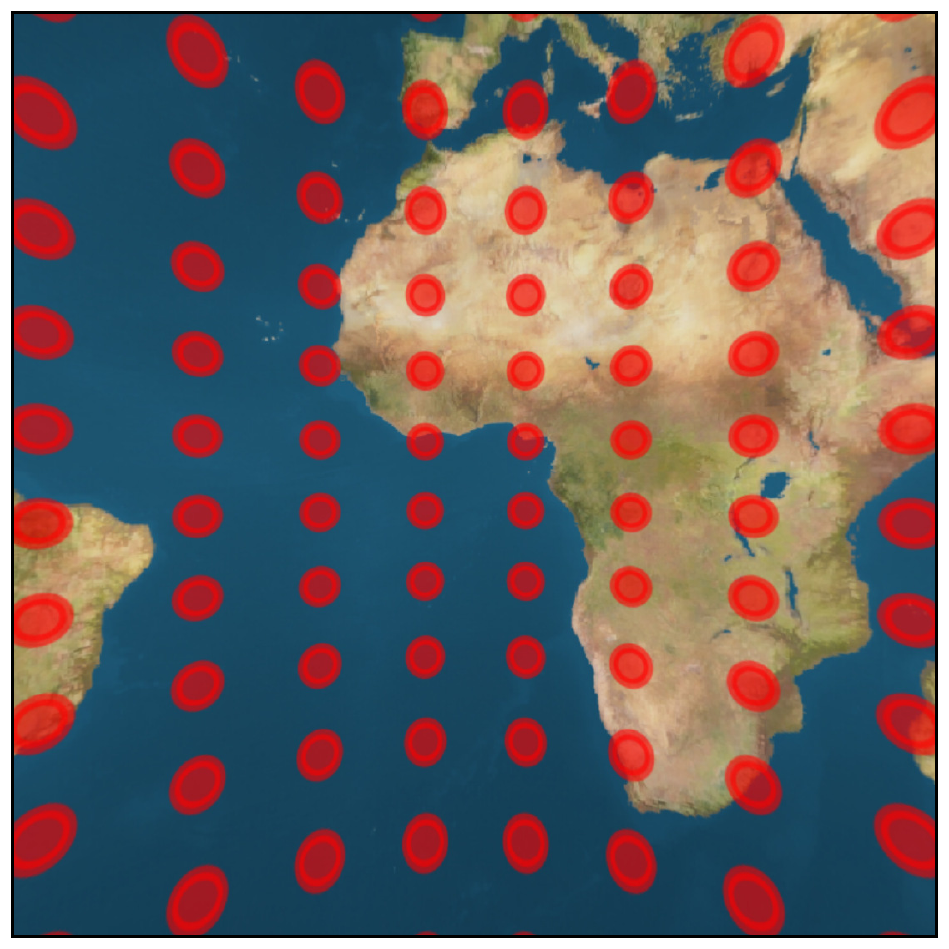}
        \caption{\label{fig:tissotgnomonic}Gnomonic}
    \end{subfigure}
    \begin{subfigure}[b]{0.48\textwidth}
        \centering
        \includegraphics[height=2.8cm, trim={0.1cm 0.1cm 0.1cm 0.1cm}, clip] {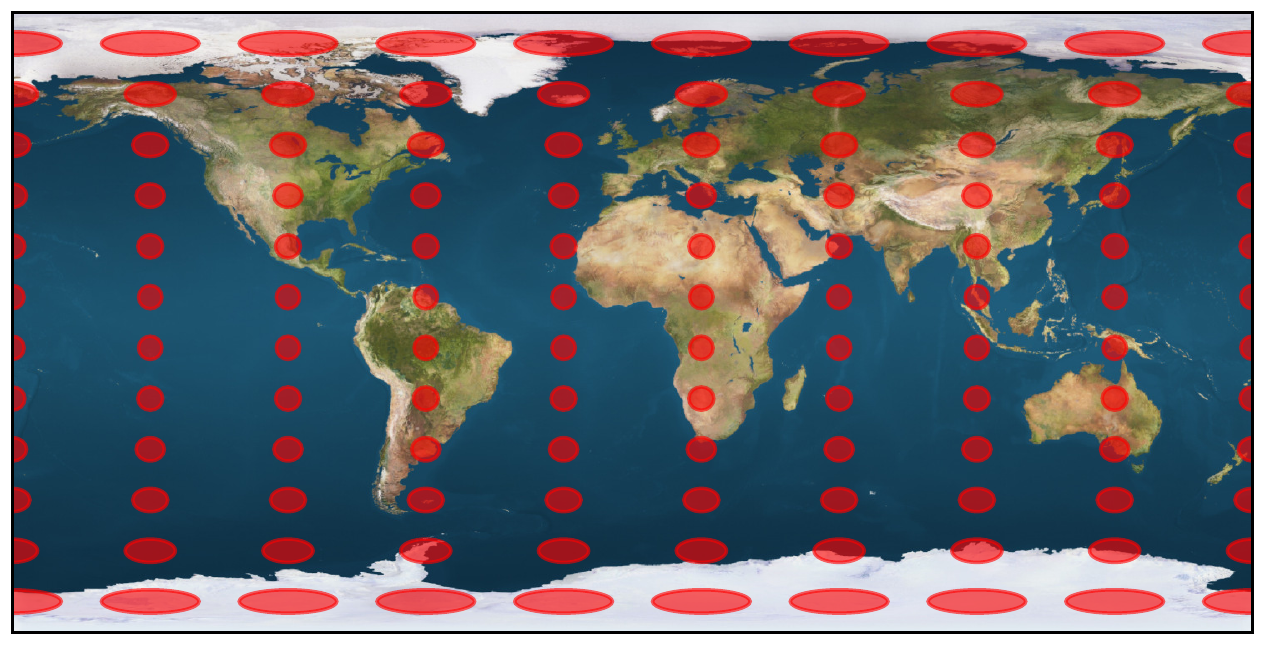}
        \caption{\label{fig:tissotequirect}Equirectangular}
    \end{subfigure}
    \begin{subfigure}[b]{0.24\textwidth}
        \centering
        \includegraphics[height=2.8cm, trim={0.1cm 0.1cm 0.1cm 0.1cm}, clip] {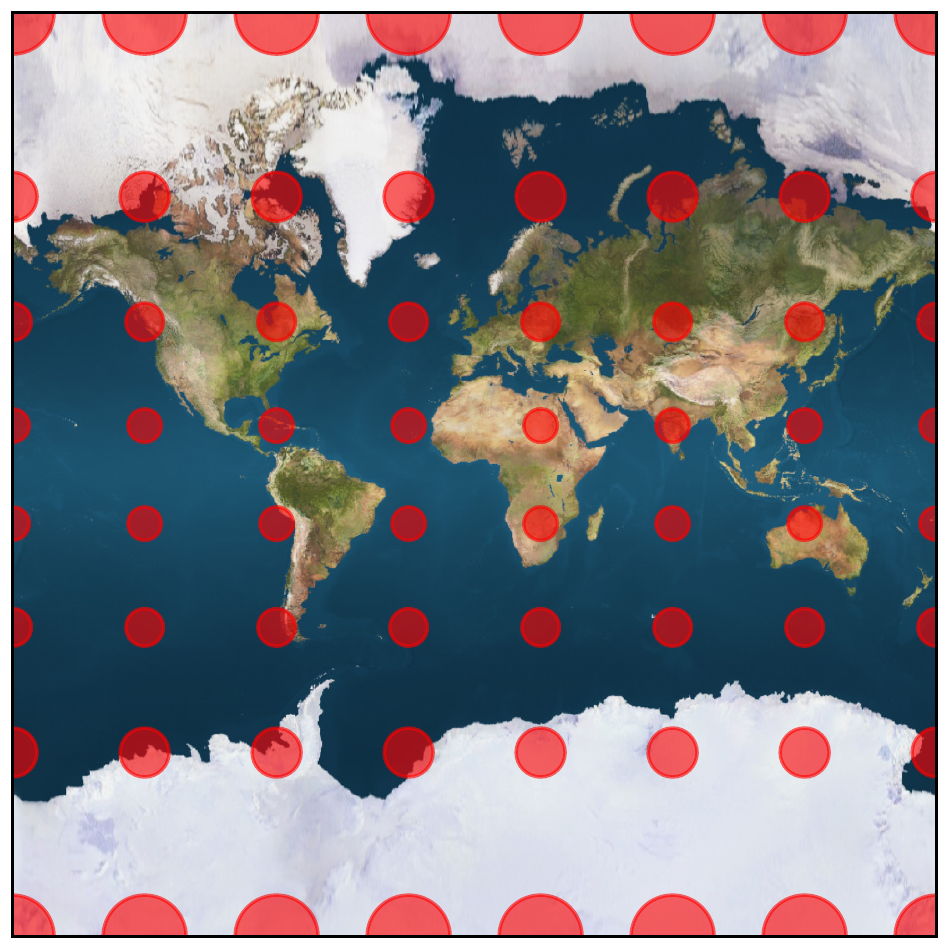}
        \caption{\label{fig:tissotmerc}Mercator}
    \end{subfigure}
    \begin{subfigure}[b]{0.66\textwidth}
        \centering
        \includegraphics[height=2.8cm, trim={0.1cm 0.1cm 0.1cm 0.1cm}, clip] {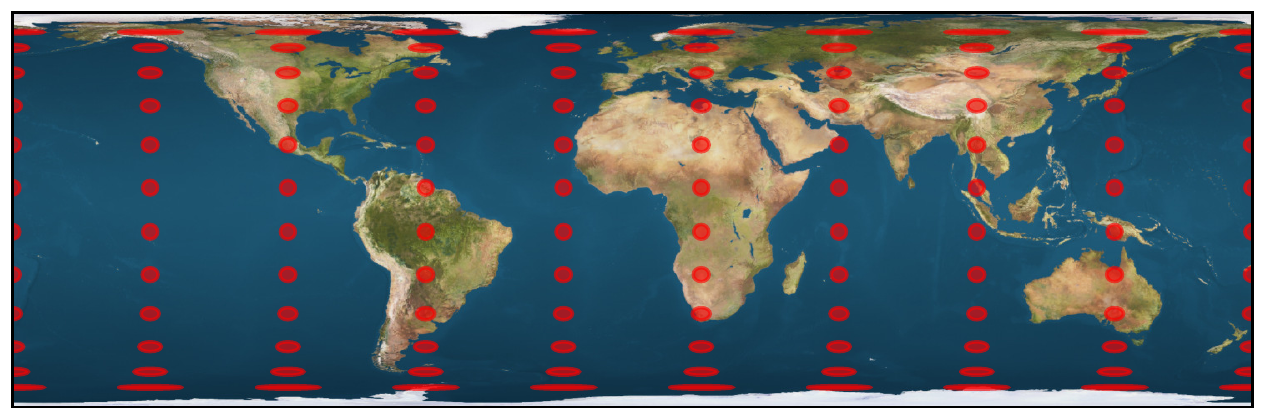}
        \caption{\label{fig:tissotgallpeters}Gall-Peters}
    \end{subfigure}
    \begin{subfigure}[b]{0.3\textwidth}
        \centering
        \includegraphics[height=2.8cm] {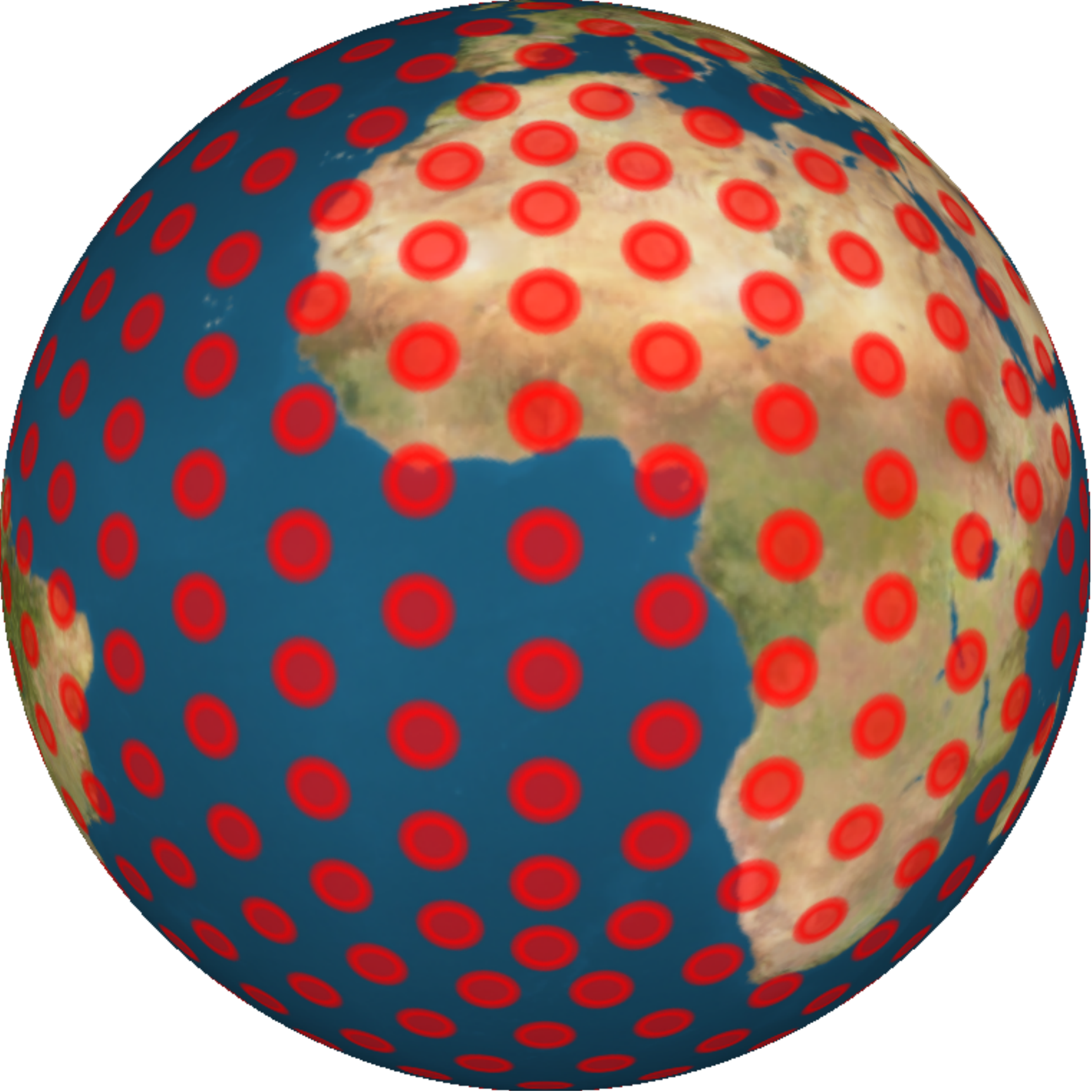}
        \caption{\label{fig:tissoticosphere}ISEA (Icosphere)}
    \end{subfigure}
\caption{Different map projections with superimposed Tissot's indicatrices to illustrate the distortion. Ellipse eccentricity shows conformality, size shows areal equality, and spacing shows equidistance. Note that the Gnomonic projection displayed only depicts a $90^{\circ} \times 90^{\circ}$ segment of the Earth, as would a face on a cube-map. Earth image from \cite{earth_image}.}
\label{fig:tissot}
\vspace{-2mm}
\end{figure*}

\subsection{Distortion in map projections}
Gauss proved with his Theorema Egregium that it is impossible to represent a sphere on a plane without some degree of distortion. The nature of this distortion is dependent on the projection model used to map the sphere to the plane. There are numerous cartographic projections that satisfy useful properties for specific applications, like navigation, but generally, there are three major projection properties to consider:
\begin{itemize}[noitemsep,topsep=2pt]
    \item \textbf{Equal area:} Preserves relative scale of objects
    \item \textbf{Conformal:} Preserves local angles
    \item \textbf{Equidistant:} Preserves distance between locations on the sphere
\end{itemize}

Resulting from Gauss's theorem, it is impossible for a projection to be both conformal and equal area, and similarly a map cannot be equidistant everywhere \cite{snyder1987map}. However, certain projections may provide one or another of these properties. A typical method for visualizing the distortion properties of different map projections is to use Tissot's indicatrices \cite{snyder1987map}, which illustrate the local distortions at points on the map. A Tissot's indicatrix represents the projection onto a map of a circle with infinitesimal radius on the surface of the sphere. The result is an ellipse where the major and minor axes relate the scale and angular distortions. By placing the circles with regular spacing on the sphere, the indicatrices also depict the distortion of distances in the projection.

Figures \ref{fig:tissotgnomonic}-\ref{fig:tissotgallpeters} show Earth represented by the gnomonic (rectilinear) projection used for faces of cube-maps, the equirectangular (plate car\'ee) projection, the common Mercator projection, and the Gall-Peters projection, respectively, with Tissot's indicatrices overlaid. The gnomonic projection is neither conformal, equal areal, nor equidistant, demonstrated by the varied eccentricity and shape of the ellipses and the spacing between them. The equirectangular projection is an equidistant projection, preserving the distance between parallels of latitude, but is neither equal-area nor conformal, so the indicatrices are neither circular nor equal in size.  The Mercator projection is conformal but not equal-area, and thus relative scale varies throughout the map. Note how Antarctica appears significantly larger than Africa, despite actually having less than 50\% of Africa's area. All indicatrices remain circular as it is conformal, but they vary in size depending on location. Finally, the Gall-Peters projection is an equal-area projection, which addresses this scale imbalance by preserving the relative size of objects in the map. This method is not conformal, though, and therefore local angles are not preserved, which can be observed by the varying eccentricities of the ellipses.

\begin{figure*}[ht]
    \centering
    \includegraphics[height=2.75cm, trim={0cm 0.1cm 0cm 0.1cm}, clip] {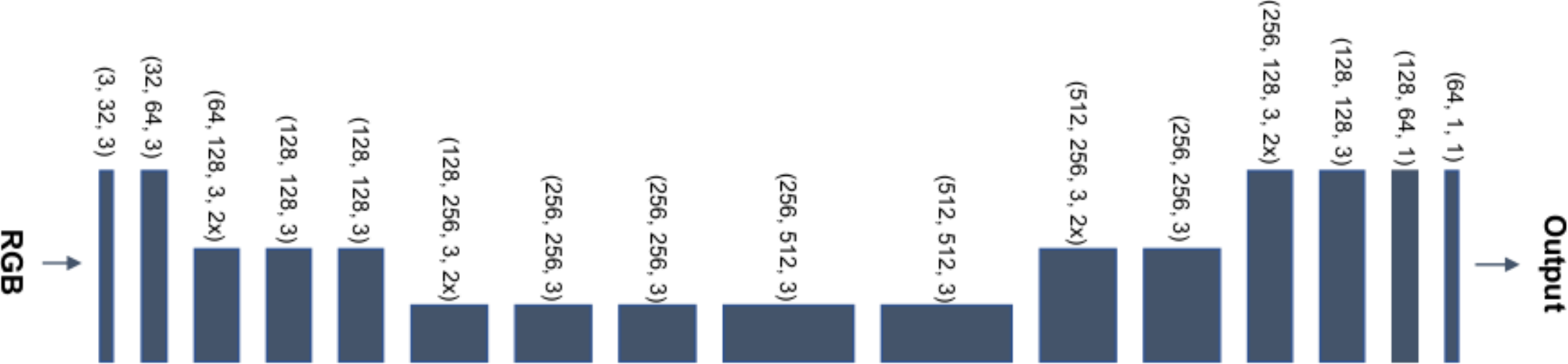}\vspace{-2mm}
\caption{Simple encoder-decoder network architecture used for our experiments. Each layer but the last is followed by an exponential linear unit activation function \cite{clevert2015fast} without batch normalization. The layers are described as \textit{(input\_channel, output\_channel, kernel\_size, up/downsampling)} and `same' padding is used for all layers.\vspace{-4mm}}
\label{fig:network}
\end{figure*}

\subsection{Relevance to convolution}\label{sec:convolution}
The 1D discrete convolution of a filter $f$ of size $K$ and a signal $g$ is given in Eq.~(\ref{eq:stdconv}):
\begin{equation}\label{eq:stdconv}
    (f * g)[n] =  \sum_{m = -\lfloor \frac{K}{2} \rfloor}^{\lfloor \frac{K}{2} \rfloor} f[m]g[n-m].
\end{equation}
The operation has two components: a sampling and a weighted summation, or more explicitly:
\begin{equation}\label{eq:stdconvsampling}
    (f * g)[n] =  \sum_{m = -\lfloor \frac{K}{2} \rfloor}^{\lfloor \frac{K}{2} \rfloor} f[m] \left(\sum_{l=-\infty}^{\infty}g[l]\delta[l-n+m]\right)
\end{equation}
where $\delta[\cdot]$ is the Dirac delta function. The Dirac delta function can be expressed as the limit of a zero-centered, isotropic Gaussian distribution as the variance goes to zero, and thus in 2D can be represented by an infinitely small circle. Hence when convolving with a spherical image, each sample is an infinitesimal circle on the surface of a sphere. As discussed previously, we can examine how this sampling is affected by distortion using Tissot's indicatrix.

From the definition of convolution, the kernel must sample the same area at each location in the data. Thus, there is an implicit assumption that the data is undistorted (i.e. the indicatrices are all perfect circles of equal area). As illustrated by Figure \ref{fig:tissotequirect}, this is not the case with equirectangular projections. Neither does it hold true for cube-maps, which use the gnomonic projection shown in Figure \ref{fig:tissotgnomonic}. These two common spherical image formats violate this distortion-free assumption, which explains why the performance of traditional CNNs on these images falls short when compared to center perspective images that typically have little to no distortion. It follows that there are two viable solutions for resolving this issue: replace the Direct delta function in Eq.~(\ref{eq:stdconvsampling}) with a bivariate, anisotropic Gaussian with location-dependent variance, or resample the image to a different projection with less distortion. The former approach would slow down convolution on spherical images, as a large variance at a sample location could require the accumulation of large regions of pixels, instead of just sampling at one. The latter approach is thus preferable. We still are blocked by the theoretical limitations of map projections, but we can select a compromise projection with the least distortion.

\subsection{Icosahedral grids}
To re-state, we desire a map projection that is: 1) nearly conformal, preserving the local shape of the data; 2) nearly equal-area, maintaining the local size of the data; and 3) nearly equidistant, allowing fixed-size convolutional filters to retain translational equivariance over the entire image. This suggests that the best projection should yield a very close planar approximation to the sphere itself. Stemming from R. Buckminster Fuller's work on discretizing the sphere \cite{fuller1982synergetics}, researchers have developed various geodesic approximations for cartographic map projections. One such projection, the icosahedral Snyder equal area (ISEA) projection \cite{snyder1992equal}, is particularly well-suited for our task of representing spherical images. It projects the data on the sphere's surface onto the planar faces of a recursively subdivided icosahedron using a triangle partitioning. This compromise projection is among the least distorted among geodesic grids \cite{kimerling1999comparing}, which satisfies our aforementioned criteria, and it is easily represented by a triangular 3D mesh. Therefore, we propose that this ISEA projection is an ideal spherical image representation for inference with CNNs.

To perform the ISEA projection, we first create a $7^{th}$ order ``icosphere,'' a regular icosahedron subdivided $7$ times using Loop subdivision \cite{loop1987smooth} with the vertices subsequently normalized to the unit sphere. We choose a $7^{th}$ order icosphere because the number of vertices ($163,842$) is most similar to the number of pixels ($131,072$) in the $256 \times 512$ equirectangular images we use for comparison. Next, we resample the image to the vertices of the icosphere using barycentric interpolation on the faces. The result of this projection is shown in Figure \ref{fig:tissoticosphere}, along with the Tissot indicatrices of deformation on the geodesic's surface. Notice how the indicatrices are all nearly equally-sized circles. This illustrates that the content deformation is very limited, and the convolution's sampling, as described in Section \ref{sec:convolution}, will be nearly unaffected. We use this representation for all spherical image inputs in our experiments.

\begin{table*}[ht]
\small
\begin{center}
\begin{tabular}{|c|c|c|c|c|c|c|c|c|c|c|c|c|c|c|c|}
\hline
Representation & Floor & Ceiling & Wall & Door & Cabinet & Rug & Window  & Curtain\\
\hline\hline
Equirect. (gnom. kernel) \cite{coors2018spherenet, tateno2018distortion} & $0.9315$ & $\mathbf{0.9710}$ & $0.8597$ & $0.6466$ & $0.6376$ & $\mathbf{0.7284}$ & $0.7012$  & $0.4703$ \\
Equirect. (equirect. kernel) & $0.9327$ & $0.9654$ & $0.8566$ & $0.6692$ & $0.6638$ & $0.7003$ & $0.7180$  & $0.4248$ \\
Icosphere (ours) & $\mathbf{0.9352}$ & $0.9703$ & $\mathbf{0.8797}$ & $\mathbf{0.6890}$ & $\mathbf{0.7037}$ & $0.6970$ & $\mathbf{0.7562}$  & $\mathbf{0.5744}$ \\
\hline
\hline
 & Sofa & Partition & Bed & Chair & Table & Shelving & Chandelier & \textit{All Classes} \\
\hline\hline
Equirect. (gnom. kernel) \cite{coors2018spherenet, tateno2018distortion} & $0.7114$ & $0.4172$ & $0.7133$ & $0.4219$ & $0.4587$ & $0.3278$ & $0.4491$  & $0.5904$ \\
Equirect. (equirect. kernel) & $0.7337$ & $0.4448$ & $0.7228$ & $0.4415$ & $\mathbf{0.5363}$ & $0.2843$ & $\mathbf{0.4566}$  & $0.5969$ \\
Icosphere (ours) & $\mathbf{0.7374}$ & $\mathbf{0.4683}$ & $\mathbf{0.7776}$ & $\mathbf{0.4375}$ & $0.5018$ & $\mathbf{0.3733}$ & $0.4472$  & $\mathbf{0.6639}$ \\
\hline
\end{tabular}
\end{center}
\vspace{-5mm}
\caption{Semantic segmentation mIOU for the different representations. Classes ordered by descending frequency.
\vspace{-4mm}
\label{tab:resultssemseg}
}
\end{table*}

\section{Semantic Segmentation Evaluation}
We evaluate the icosphere representation using semantic segmentation and demonstrate that limiting distortion in the representation spawns significant accuracy improvements.

\subsection{Dataset}
For our experiments, we use the SUMO \cite{sumo} dataset, a derivative of SunCG \cite{suncg} containing $58,631$ RGB-D cube-maps representing synthetic spherical image captures. We divide the SUMO dataset into 80\% training, 10\% validation, and 10\% testing splits and pre-process the data by resampling the images onto an $7^{th}$ order icosphere. An example of this resampling is displayed with a super-imposed convolutional kernel in Figure \ref{fig:teaser}. As the dataset contains $132$ semantic classes, some with high inter-class similarity (e.g. `bunk bed' and `baby bed'), we aggregated these similar classes to balance the data for these experiments. There is still a very high variance in class frequency even after aggregation, so we only evaluate performance on the $15$ most common classes. For exact comparison with existing work, we render the predictions on the icosphere back into an equirectangular image to assess per-pixel accuracy.
\vspace{-2mm}
\subsection{CNN Training}
We follow the method of Coors \etal \cite{coors2018spherenet} and Tateno \etal \cite{tateno2018distortion} to sample the spherical data using a gnomonic projection of the planar filter. However, as our data is no longer in a 2D image, we project the planar filter onto the icosphere's surface and sample from the faces using barycentric interpolation. We set the filter resolution to the mean distance between adjacent vertices on the icosphere, to maintain analogy with the notion of resolution of the planar filter being set to the distance between adjacent pixels. During convolution, we apply this filter at each vertex of the icosphere and sample from the faces using barycentric interpolation of the data at the $3$ vertices. For up- and down-sampling, we leverage different subdivision orders, which, similar to pixels in 2D images, vary roughly by a factor of 4 in the number of vertices.  We compare convolving with our icosphere representation to the method of \cite{coors2018spherenet, tateno2018distortion}, which convolves with equirectangular images. We use the encoder-decoder architecture given in Figure \ref{fig:network} for both methods. We train for 10 epochs using Adam \cite{kingma2014adam} with an initial learning rate of $10^{-4}$, reduced by half every 3 epochs. Our training criterion is the cross-entropy loss weighted by inverse class frequency.

\subsection{Results and Discussion}
\begin{figure}
    \centering
        \begin{subfigure}[b]{0.23\textwidth}
        \centering
        \includegraphics[width=\textwidth, trim={0.5cm 1.5cm 1.5cm 2.7cm}, clip] {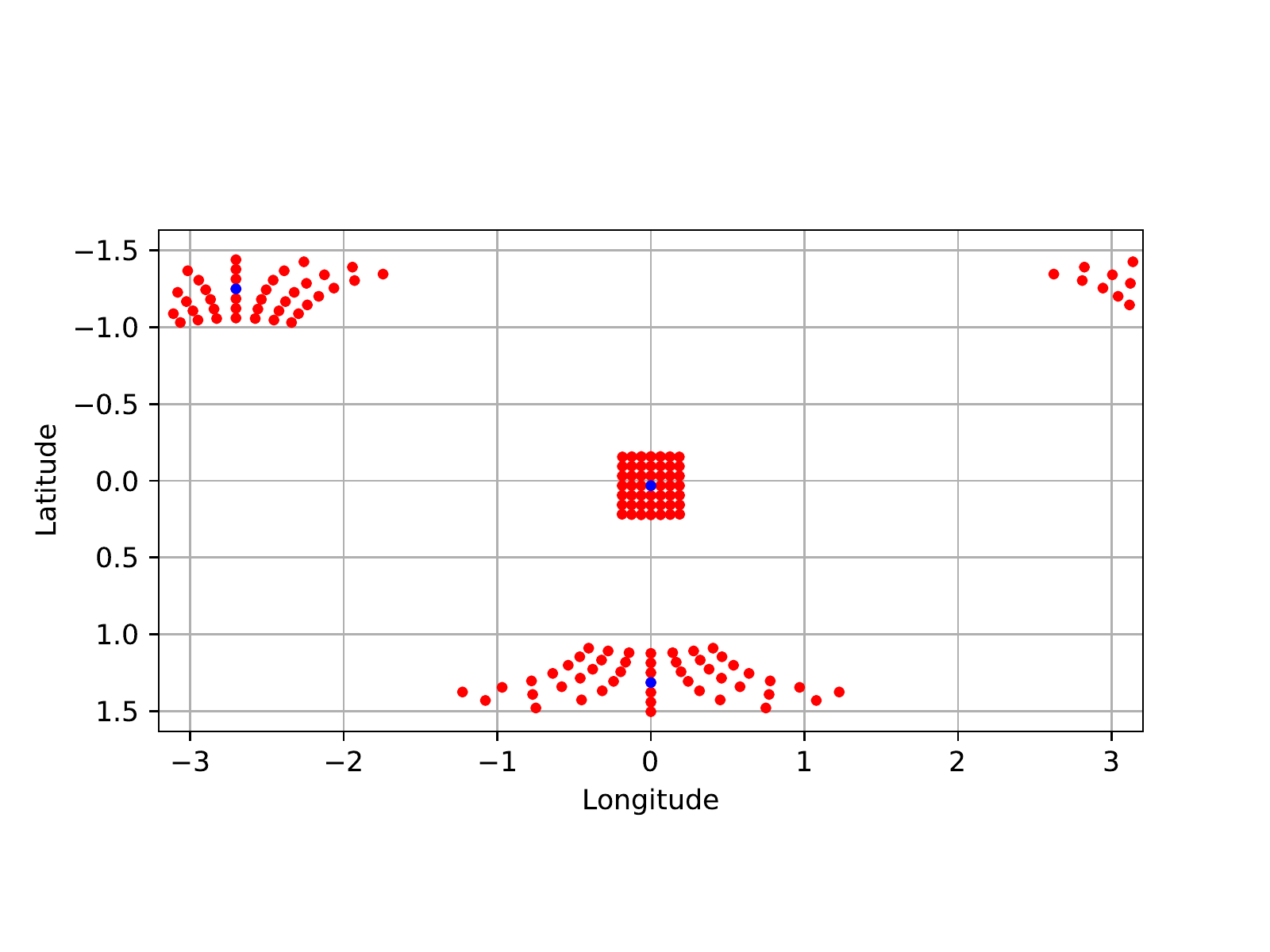}\vspace{-2mm}
        \caption{Gnomonic\vspace{-2mm}}
\end{subfigure}
    \centering
    \begin{subfigure}[b]{0.23\textwidth}
        \centering
        \includegraphics[width=\textwidth, trim={0.5cm 1.5cm 1.5cm 2.7cm}, clip] {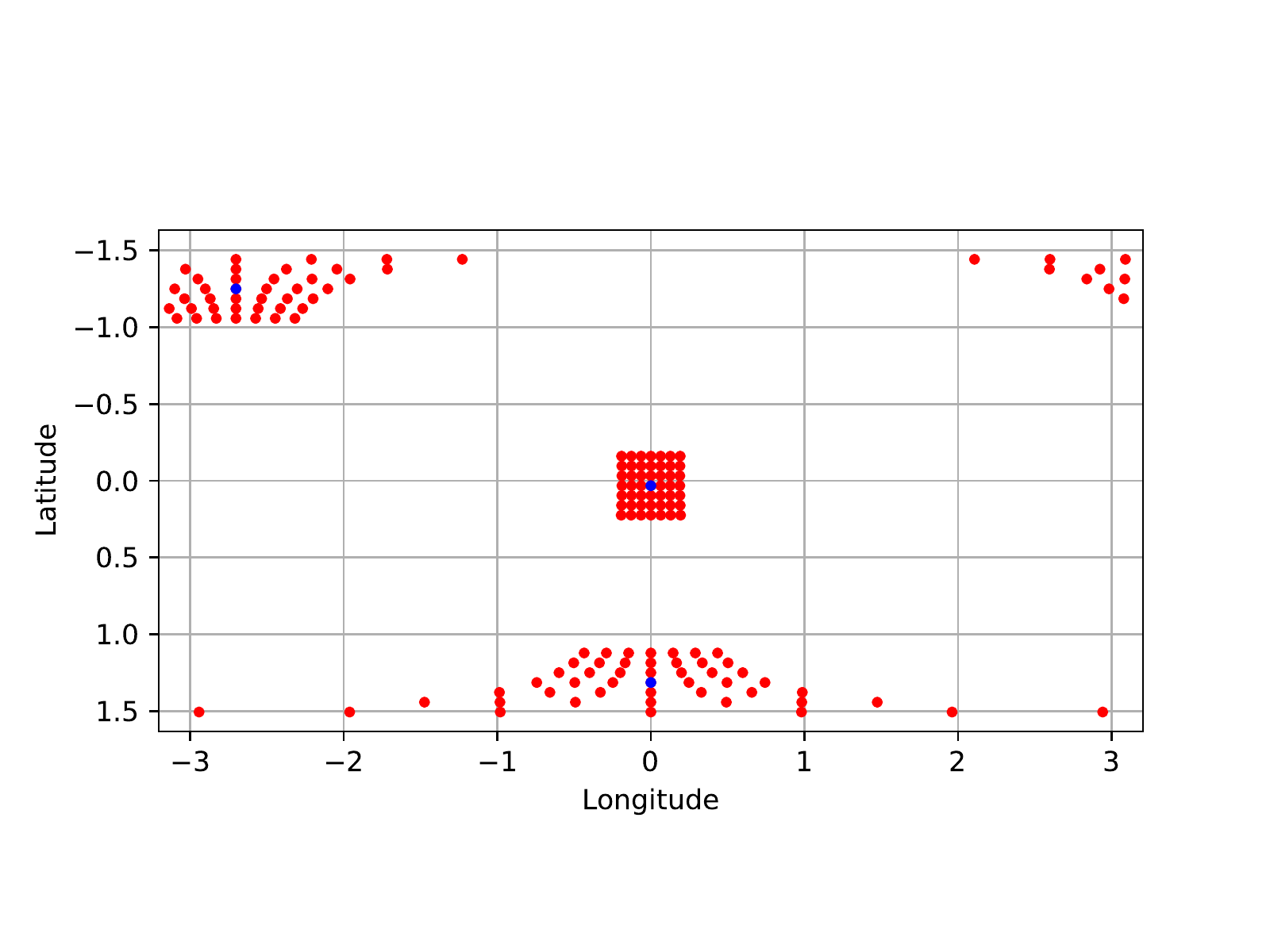}\vspace{-2mm}
        \caption{Equirectangular\vspace{-2mm}}
    \end{subfigure}
\caption{\label{fig:samplingcomp}Comparing convolutional kernel sampling patterns using gnomonic and equirectangular projections.
\vspace{-4mm}}
\end{figure}

Results of our semantic segmentation experiments are given in Table \ref{tab:resultssemseg} as the per-class mean intersection-over-union (mIOU). While certain classes perform marginally better in the equirectangular format, we see an $12.6\%$ improvement in the overall mIOU compared to \cite{coors2018spherenet, tateno2018distortion} simply by resampling the image. There are two reasons for this jump. The first is that the gnomonic projection used for the kernel in \cite{coors2018spherenet, tateno2018distortion} is not optimal for modeling the distortion in an equirectangular image. In fact, an equirectangular projection of the kernel would be a more appropriate, as it would correctly adapt the sampling locations to equirectangular distortion. The difference in the sampling patterns is slight, though, as shown in Figure \ref{fig:samplingcomp}, and the quantitative result is only slightly better (second row in Table \ref{tab:resultssemseg}). It is worth mentioning that the gnomonic kernel does make sense for sampling from the icosphere, as it represents the projection of a tangent plane onto a sphere. In our case, the planar filter is tangent to our icosahedral spherical approximation. Nevertheless, the more important consideration is that neither the gnomonic nor equirectangular filter projections correct for the information imbalance caused by the distortion. The filter projection method of \cite{coors2018spherenet, tateno2018distortion} dynamically adjusts where the kernel samples, but it only accounts for the conformality and equidistance properties of distortion. It does not address the areal inequality. Unequal area projections disrupt the distribution of information throughout the image. For example, all of the pixels along the top row in an equirectangular images are essentially redundant, having all been sampled from the pole. That whole row carries the same information capacity of a single pixel at the equator. For convolutions, this means that the amount of information aggregated at each pixel varies throughout the image. The ISEA projection, being nearly equal-area, corrects for this, ensuring that predictions at each vertex of the icosphere  are made by accumulating equivalent quantities of information. This translates to better inference. It is also worth noting that the objects for which the equirectangular representation perform marginally better are typically found at the top (ceiling, chandelier) or bottom (rug) of an image. The equirectangular format's pixel redundancy increases the number of training samples for these objects, which would explain the better performance at test time.
\vspace{-2mm}
\section{Conclusion}
Borrowing from the field of cartography, we have identified three key principles of local distortion that impact our ability to convolve on spherical data. Applying these principles to our choice of data representation, we have proposed the ISEA projection on an icosphere as an ideal spherical image representation. We demonstrated that by significantly limiting distortion, this representation provides a $12.6\%$ boost in semantic segmentation performance.

{\small
\bibliographystyle{ieee_fullname}

}

\newpage
\begin{center}
\Large\textbf{Convolutions on Spherical Images: Supplementary Material}
\end{center}

\section{Qualitative Results}
\begin{figure*}
    \centering

    \raisebox{23pt}{\parbox[b]{.03\textwidth}{1)}}%
    \begin{subfigure}[b]{0.23\textwidth}
        \centering
        \includegraphics[width=\textwidth]{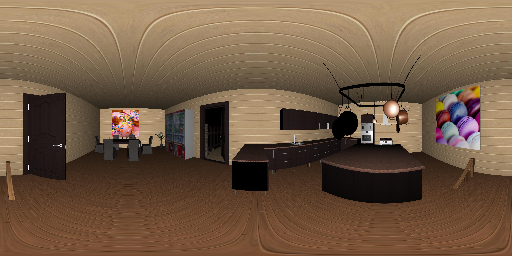}
    \end{subfigure}
    ~
    \centering
    \begin{subfigure}[b]{0.23\textwidth}
        \centering
        \includegraphics[width=\textwidth]{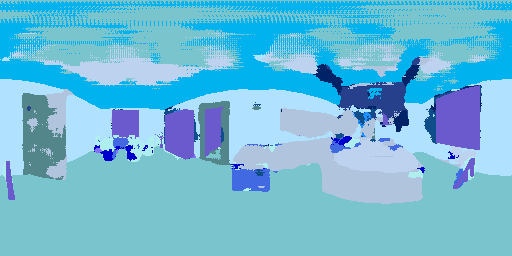}
    \end{subfigure}
    ~
    \centering
    \begin{subfigure}[b]{0.23\textwidth}
        \centering
        \includegraphics[width=\textwidth]{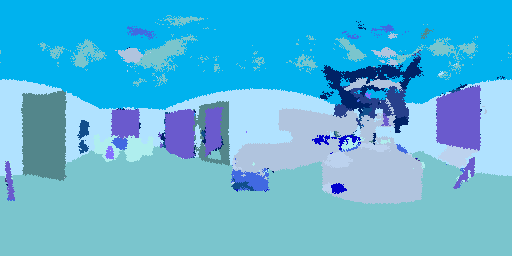}
    \end{subfigure}
    ~
    \centering
    \begin{subfigure}[b]{0.23\textwidth}
        \centering
        \includegraphics[width=\textwidth]{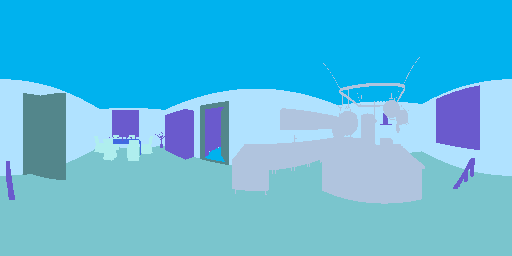}
    \end{subfigure}\\

    \raisebox{23pt}{\parbox[b]{.03\textwidth}{2)}}%
    \begin{subfigure}[b]{0.23\textwidth}
        \centering
        \includegraphics[width=\textwidth]{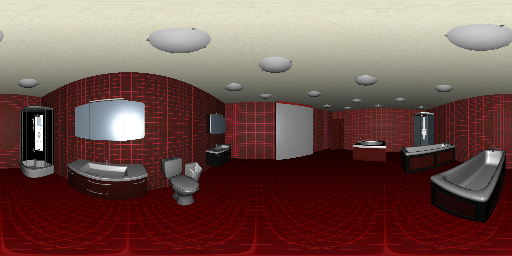}
    \end{subfigure}
    ~
    \centering
    \begin{subfigure}[b]{0.23\textwidth}
        \centering
        \includegraphics[width=\textwidth]{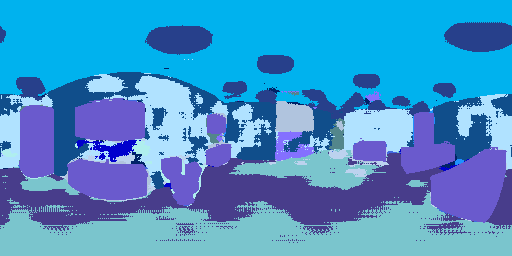}
    \end{subfigure}
    ~
    \centering
    \begin{subfigure}[b]{0.23\textwidth}
        \centering
        \includegraphics[width=\textwidth]{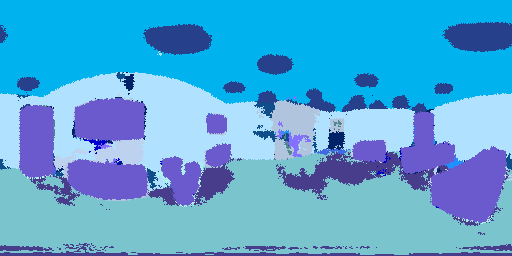}
    \end{subfigure}
    ~
    \centering
    \begin{subfigure}[b]{0.23\textwidth}
        \centering
        \includegraphics[width=\textwidth]{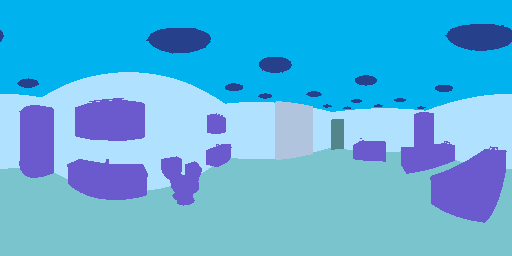}
    \end{subfigure}\\

    \raisebox{23pt}{\parbox[b]{.03\textwidth}{3)}}%
    \begin{subfigure}[b]{0.23\textwidth}
        \centering
        \includegraphics[width=\textwidth]{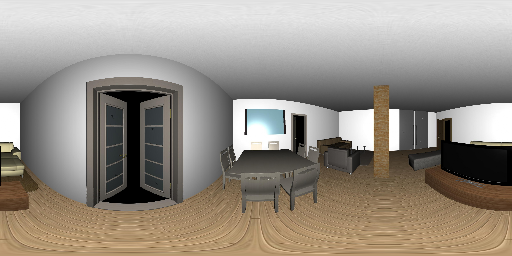}
    \end{subfigure}
    ~
    \centering
    \begin{subfigure}[b]{0.23\textwidth}
        \centering
        \includegraphics[width=\textwidth]{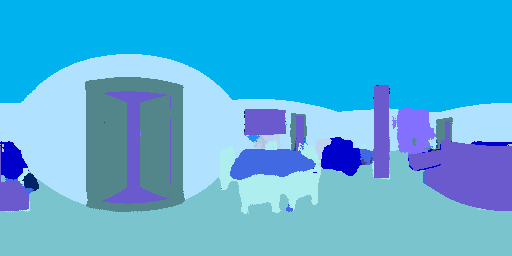}
    \end{subfigure}
    ~
    \centering
    \begin{subfigure}[b]{0.23\textwidth}
        \centering
        \includegraphics[width=\textwidth]{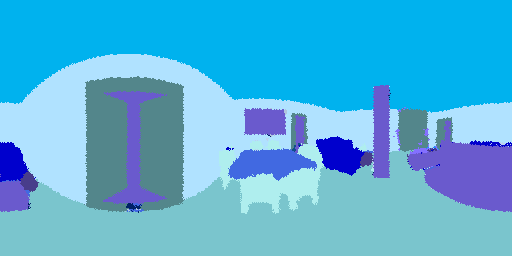}
    \end{subfigure}
    ~
    \centering
    \begin{subfigure}[b]{0.23\textwidth}
        \centering
        \includegraphics[width=\textwidth]{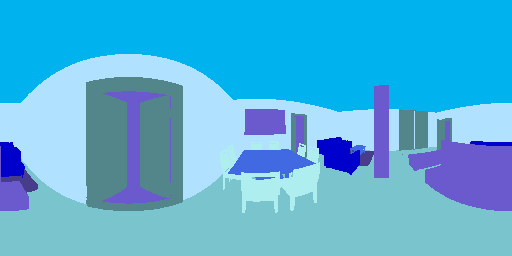}
    \end{subfigure}\\

    \raisebox{23pt}{\parbox[b]{.03\textwidth}{4)\label{fig:rowfive}}}%
    \begin{subfigure}[b]{0.23\textwidth}
        \centering
        \includegraphics[width=\textwidth]{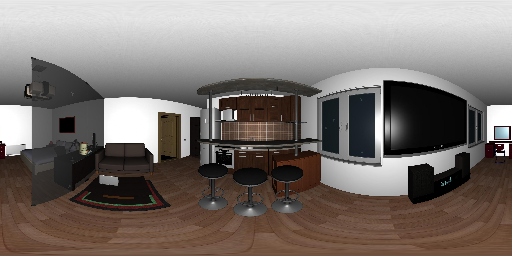}
    \end{subfigure}
    ~
    \centering
    \begin{subfigure}[b]{0.23\textwidth}
        \centering
        \includegraphics[width=\textwidth]{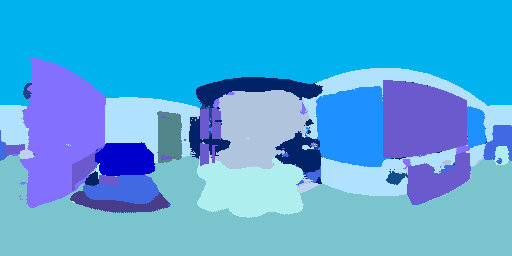}
    \end{subfigure}
    ~
    \centering
    \begin{subfigure}[b]{0.23\textwidth}
        \centering
        \includegraphics[width=\textwidth]{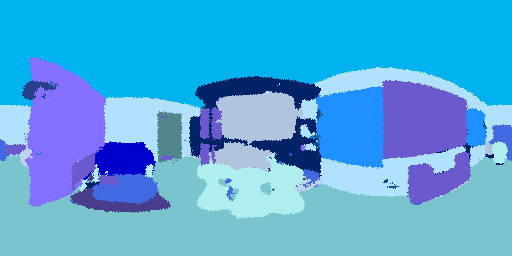}
    \end{subfigure}
    ~
    \centering
    \begin{subfigure}[b]{0.23\textwidth}
        \centering
        \includegraphics[width=\textwidth]{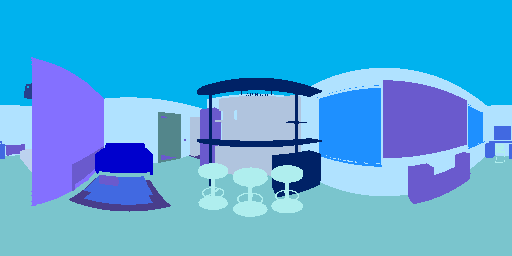}
    \end{subfigure}\\

    \raisebox{23pt}{\parbox[b]{.03\textwidth}{5)}}%
    \begin{subfigure}[b]{0.23\textwidth}
        \centering
        \includegraphics[width=\textwidth]{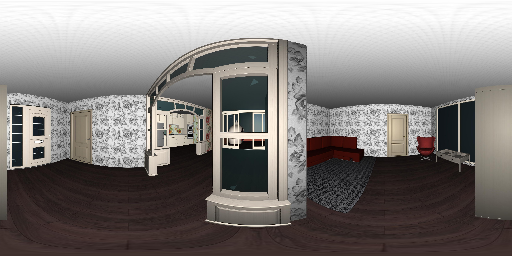}
    \end{subfigure}
    ~
    \centering
    \begin{subfigure}[b]{0.23\textwidth}
        \centering
        \includegraphics[width=\textwidth]{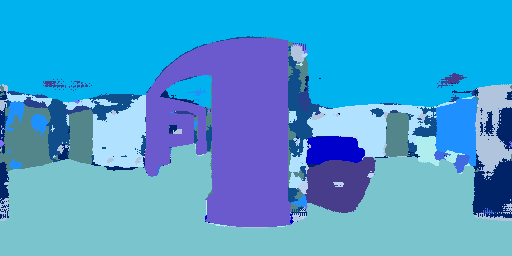}
    \end{subfigure}
    ~
    \centering
    \begin{subfigure}[b]{0.23\textwidth}
        \centering
        \includegraphics[width=\textwidth]{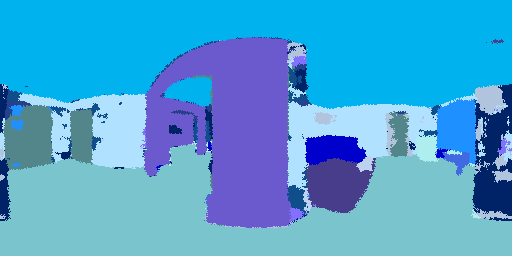}
    \end{subfigure}
    ~
    \centering
    \begin{subfigure}[b]{0.23\textwidth}
        \centering
        \includegraphics[width=\textwidth]{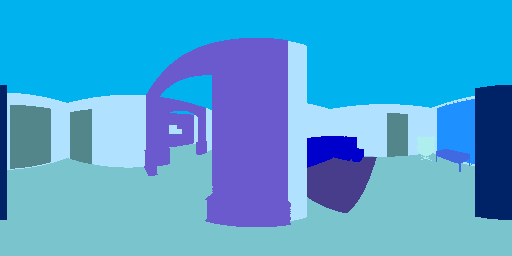}
    \end{subfigure}\\

    \raisebox{23pt}{\parbox[b]{.03\textwidth}{6)}}%
    \begin{subfigure}[b]{0.23\textwidth}
        \centering
        \includegraphics[width=\textwidth]{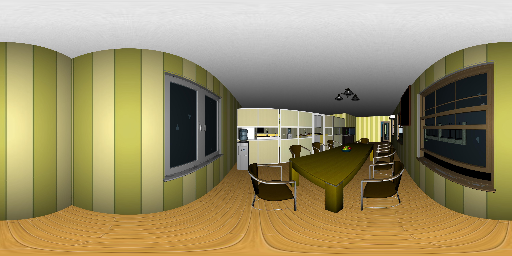}
    \end{subfigure}
    ~
    \centering
    \begin{subfigure}[b]{0.23\textwidth}
        \centering
        \includegraphics[width=\textwidth]{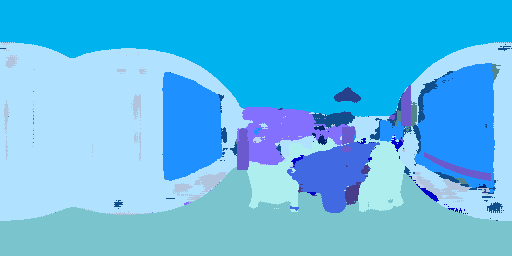}
    \end{subfigure}
    ~
    \centering
    \begin{subfigure}[b]{0.23\textwidth}
        \centering
        \includegraphics[width=\textwidth]{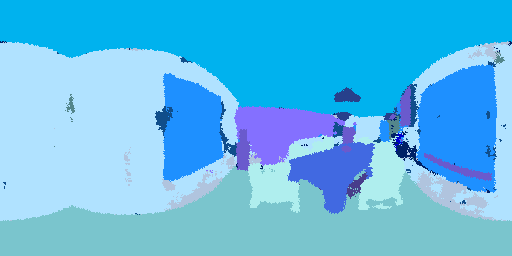}
    \end{subfigure}
    ~
    \centering
    \begin{subfigure}[b]{0.23\textwidth}
        \centering
        \includegraphics[width=\textwidth]{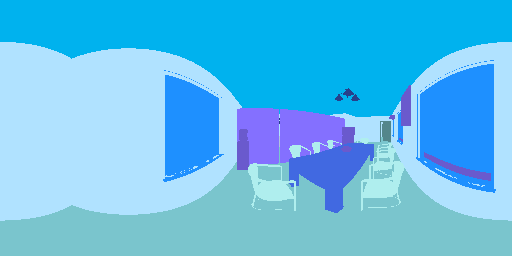}
    \end{subfigure}\\

    \raisebox{38pt}{\parbox[b]{.03\textwidth}{7)}}%
    \begin{subfigure}[b]{0.23\textwidth}
        \centering
        \includegraphics[width=\textwidth]{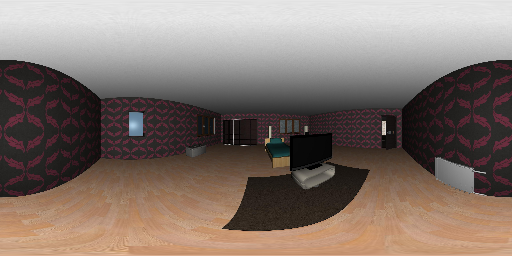}
        \caption{Input}
    \end{subfigure}
    ~
    \centering
    \begin{subfigure}[b]{0.23\textwidth}
        \centering
        \includegraphics[width=\textwidth]{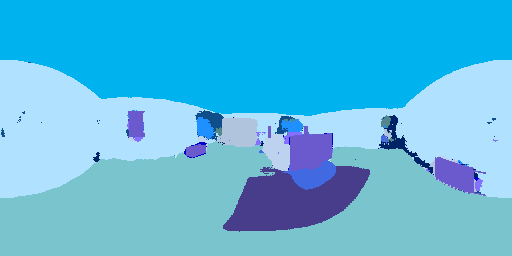}
        \caption{Equirectangular Image}
    \end{subfigure}
    ~
    \centering
    \begin{subfigure}[b]{0.23\textwidth}
        \centering
        \includegraphics[width=\textwidth]{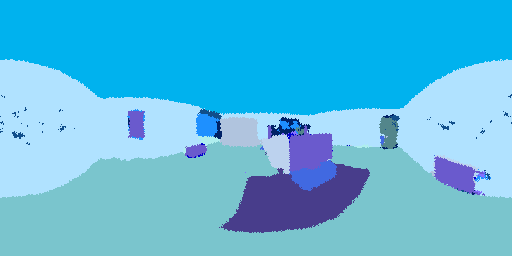}
        \caption{Icosphere}
    \end{subfigure}
    ~
    \centering
    \begin{subfigure}[b]{0.23\textwidth}
        \centering
        \includegraphics[width=\textwidth]{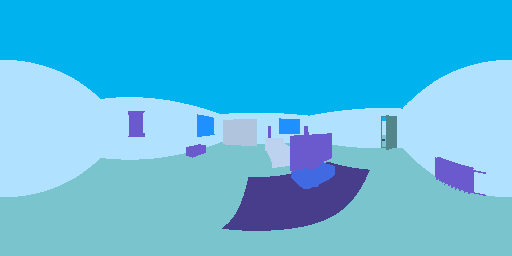}
        \caption{Ground Truth}
    \end{subfigure}\\

\caption{Sample semantic segmentation results. In column (b) are the results of applying the method of \cite{coors2018spherenet, tateno2018distortion} to equirectangular image representations of spherical data. In column (c) are the results of our method using the ISEA representation on the icosphere. The icosphere results are resampled back to an equirectangular projection for visualization here. Note that the visualizations in column (c) have pixelated edges due to the use of nearest-vertex interpolation in the resampling from the icosphere. This is an artifact of the rendering, not a result of the predictions.}
\label{fig:examples}
\end{figure*}

Figure \ref{fig:examples} shows a qualitative comparison between semantic segmentation results obtained using the equirectangular image representation and our proposed ISEA representation on the icosphere. As this work is focused on examining the methodology, the results are not post-processed in any way, and we note that the architecture and loss we use are not state-of-the-art. However, the representation of the data is network-agnostic, and we expect the outcome to be the same regardless of the choice of network architecture and training routine.

The quantitative results in Table \ref{tab:resultssemseg} of the main paper illustrate that the ISEA representation improves segmentation quality. This is borne out qualitatively as well. In Figure \ref{fig:examples}, rows 1 and 2 provide clear examples of this effect. The improvement comes from both more accurate pixel classification, as with the door in the center-right of the image in row 3, and more precise segmentation coverage, as with the stools in the center of the image in row 4. Rows 5 and 6 show examples where our representation has provided a better quantitative result, but the actual segmentations are not noticeably better than with the equirectangular image representation. Instead, we seem to trade some mistakes for others. The last row of samples demonstrates that the equirectangular image representation can achieve similar performance to the icosphere representation. In fact, there are many examples where differences in the qualitative results are indiscernible. Nonetheless, we conclude that there is a clear and demonstrable benefit to using the ISEA representation, justified by both the quantitative metrics and the qualitative comparison given here.

\end{document}